\documentclass[pdflatex,sn-basic]{sn-jnl}

\usepackage{graphicx}
\usepackage{amsmath,amssymb,amsfonts}
\usepackage{booktabs}
\usepackage{multirow}
\usepackage{tabularx}
\usepackage{array}
\usepackage[section]{placeins}
\usepackage{tikz}
\usetikzlibrary{arrows.meta,positioning}

\newcolumntype{Y}{>{\raggedright\arraybackslash}X}

\articletype{Review Article}

\title[Hadith computational science]{Hadith computational science in the age of large language models: a critical narrative review}

\author[1]{\fnm{Md. Ashraful} \sur{Haque}}\email{ashraful.haque@gtaf.org}

\author*[1,2]{\fnm{Riasat} \sur{Islam}}\email{riasat.islam@gtaf.org}\email{riasat.islam@qmul.ac.uk}

\affil[1]{\orgname{Greentech Apps Foundation UK}, \orgaddress{\country{United Kingdom}}}

\affil*[2]{\orgname{Queen Mary University of London}, \orgaddress{\city{London}, \country{United Kingdom}}}

\abstract{We examine how hadith computational science is being reshaped by transformer models, retrieval-grounded pipelines, and large language models (LLMs). Recent reviews document growth in the literature, but they do not yet provide a critical account of which advances are methodologically robust, which remain benchmark-bound, and which unresolved problems still limit scholarly use. We address this gap through a critical narrative review that combines critique of existing reviews, paper-level appraisal of representative original studies, and synthesis of Islamic scholar and domain-expert perspectives on authenticity, authority, and responsible use. We find uneven progress. Data resources have expanded, segmentation tasks have matured, narrator and source-verification problems are better formalized, and LLM-assisted workflows now support corpus-scale enrichment, multilingual access, and grounded evaluation. At the same time, progress remains constrained by narrow corpora, weak benchmark comparability, synthetic-to-real transfer gaps, narrator identity resolution, preprocessing fragility, limited reproducibility, and sparse expert-grounded validation. We show that important gaps lie beyond dominant benchmarks: non-canonical and obscure corpora, commentary and explanatory literature, cross-source links with Qur'an and seerah, and fiqh-facing evidence support. We argue that hadith computation should be assessed less as isolated model performance than as an evidence infrastructure problem requiring knowledge integration, provenance, and expert supervision. On this basis, we define a research agenda for making the field methodologically stronger and more useful to Islamic scholarship.}


\begin{document}

\maketitle

\section{Introduction}\label{sec:introduction}

\citet{azmi2019computational} provided the clearest early map of hadith computation. Their survey described a field shaped mainly by rule-based systems, classical machine learning, small curated corpora, and task-specific pipelines. It documented important work on segmentation, narrator extraction, ontology construction, retrieval, and authentication support. It also captured the field before transformer-based NLP and generative models reshaped expectations about scale, representation, and end-to-end language processing.

This review addresses a field now operating in a markedly different technical environment. Transformer-based models, retrieval-grounded systems, and large language models have changed the scale of modeling, the structure of evaluation, and the plausibility of corpus-wide workflows. Hadith computation has changed with them, but unevenly. Some tasks, especially isnad-matn separation, that is, separating the chain of transmission from the report content, and corpus enrichment, have benefited from better data and modern modeling \citep{altammami2020isnad, muther2020automatic, asgari2025rezwan}. Other problems, including narrator disambiguation, cross-collection generalization, biographical grounding, and trustworthy validation, remain difficult in ways that larger models alone have not removed \citep{mahmoud2022arsanad, mghari2022sanadset, alshuhayeb2025noor}.

Recent review papers document a growing literature \citep{sulistio2023utilization, hakak2022recent, azwar2025bibliometric, azwar2025prisma}. These reviews are useful, but they leave open a different interpretive question: which original studies provide durable evidence of methodological progress, which claims remain local to particular benchmarks, and which weaknesses continue to limit downstream scholarly use. Most also give limited attention to Islamic scholar and domain-expert perspectives, even though those perspectives become more important as AI systems move closer to religiously sensitive tasks \citep{akbar2024digitalturn, elmahjub2023islamicethics}.

What remains needed is a critical synthesis that links contemporary technical developments to the scholarly tasks they are meant to support. We therefore address five linked questions. Which methodological changes appear most important in the transformer and LLM era? Which contemporary studies represent substantive advances rather than benchmark-local gains? Why do existing review papers still leave an interpretive gap? Which structural limitations continue to constrain the field, especially beyond the six canonical books and beyond base hadith texts alone? And how should Islamic scholar and domain-expert perspectives inform the next research phase?

We adopt a critical narrative review design. The field remains heterogeneous in data, methods, venues, and evaluation practice, so exhaustive counting alone can flatten the differences that matter most. Our aim is to distinguish durable advances from benchmark-local gains, identify where the field appears to be maturing, and explain why the next phase of hadith computation will depend as much on epistemic grounding and expert-aligned evaluation as on model scale.

\section{Limits of existing reviews}\label{sec:positioning}

Our aim is not to fill an absence of review literature, but to address a mismatch between the kinds of reviews currently available and the kinds of questions the field now demands. Existing reviews are useful, but they answer adjacent questions rather than the one we center here.

\begin{table}[!htbp]
\caption{How our review differs from recent review literature on hadith computation and digital hadith studies.}
\label{tab:review_positioning}
\small
\begin{tabularx}{\textwidth}{p{2.4cm}Y Y}
\toprule
\textbf{Review} & \textbf{Useful contribution} & \textbf{Critical limitation relative to our review} \\
\midrule
Luthfi et al. \citeyearpar{luthfi2018digital} & Early review of digital hadith authentication literature. & Valuable historical baseline for one application area, but narrow in scope and fully pre-transformer. \\
\addlinespace
Azmi et al. \citeyearpar{azmi2019computational} & Foundational broad survey; establishes the field's earlier technical baseline. & Necessary starting point, but it precedes the transformer- and LLM-shaped research landscape examined here. \\
\addlinespace
Hakak et al. \citeyearpar{hakak2022recent} & Strong summary of digital hadith authentication advances, challenges, and future directions. & Authentication-focused; does not reinterpret the wider computational landscape or the LLM shift across tasks. \\
\addlinespace
Sulistio et al. \citeyearpar{sulistio2023utilization} & Systematic overview of machine-learning use in hadith studies. & More technology inventory than paper-level critique; says little about benchmark comparability, task maturity, or which original studies changed the field. \\
\addlinespace
Azwar et al. \citeyearpar{azwar2025bibliometric} & Maps publication growth, authors, institutions, and topic trends. & Bibliometric evidence is weak evidence of technical maturity; publication counts do not show whether core problems are being solved. \\
\addlinespace
Azwar and Usman \citeyearpar{azwar2025prisma} & Frames AI in hadith studies through a PRISMA-based systematic review. & Useful for trend synthesis, but still high-level and descriptive; it does not provide sustained critical appraisal of representative original studies. \\
\midrule
Our review & Combines critique of review papers, critical evaluation of representative original studies, and Islamic scholar/domain-expert perspectives. & Addresses the gap left by reviews that map activity but do not evaluate evidentiary strength, task maturity, and the implications of contemporary AI for hadith scholarship. \\
\bottomrule
\end{tabularx}
\end{table}

The main limitation of recent review papers lies less in inaccuracy than in analytical level. Bibliometric and systematic reviews are well suited to showing growth, topical clustering, and method labels. They are less well suited to determining whether a reported advance is benchmark-specific, whether a resource is reusable, whether a task addresses a genuine scholarly bottleneck or only a proxy, and whether the outputs are trustworthy enough for a high-stakes religious text domain.

A critical review should not merely list domain applications. It should evaluate what techniques, model classes, and evaluation designs have actually achieved, where the evidence is thin, and why some claims travel poorly outside narrow benchmark settings. In our view, much of the current hadith review literature remains too descriptive at that level.

This limitation becomes more visible in the transformer and LLM era. Once the field began to adopt newer model families and pipeline-scale workflows, it could no longer be adequately summarized by listing additional studies and architectures. A useful review must therefore identify substantive methodological change rather than publication growth alone, move from descriptive aggregation to critical appraisal of representative original papers, explain why existing technical progress still falls short of robust scholarly use, and account for the fact that Islamic scholars and domain experts increasingly frame the central issues not only as performance questions, but also as questions of authority, authenticity, explainability, and ethical governance \citep{akbar2024digitalturn, nawi2021muslimexperts, elmahjub2023islamicethics}.

We do not present this review as a substitute for bibliometric or PRISMA-style work. Instead, we use a different review logic to ask a different question: which parts of the literature provide the strongest evidence of methodological progress, what kind of evidence those studies actually provide, and where technical progress still fails to support robust scholarly use.
\FloatBarrier

\section{Review design and limitations}\label{sec:design}

\subsection{Search and scope}

We follow a critical narrative review design, but we make the assembly process explicit. Searches were conducted iteratively during late 2025 and early 2026 across Google Scholar, Scopus, ACL Anthology, SpringerLink, ScienceDirect, arXiv, and manually curated domain repositories. Query families combined hadith terms such as \emph{hadith}, \emph{sanad}/\emph{isnad} (chain of transmission), \emph{matn} (report content), \emph{sharh} (commentary), and \emph{takhrij} (source tracing and authentication referencing) with AI terms such as \emph{NLP}, \emph{machine learning}, \emph{deep learning}, \emph{transformer}, \emph{large language model}, \emph{knowledge graph}, \emph{question answering}, \emph{retrieval}, and \emph{authentication}. Forward and backward citation tracing was then used to consolidate the working list.

The review database assembled for this study contained 42 items at the initial shortlist stage and 32 records after scope refinement. The refined inventory included technical studies, dataset or corpus papers, review papers, and a smaller set of contextual works used to interpret corpus infrastructure and governance questions. We place particular emphasis on work associated with the transformer and LLM era while retaining earlier studies that define the field's technical baseline or remain methodologically instructive.

\subsection{Inclusion logic and study types}

We included a record when it met at least one of three conditions. First, it introduced or evaluated computational methods directly on hadith or closely related classical Arabic religious text tasks. Second, it created, documented, or benchmarked resources that materially affect hadith computation, such as corpora, segmentation datasets, narrator datasets, or knowledge-grounded evaluation assets. Third, it offered review-level or scholar-oriented reflections that change how the field should be interpreted, especially regarding authenticity, authority, or responsible AI use in Islamic scholarship. We excluded purely theological discussions without computational relevance, as well as generic Arabic NLP papers that did not materially illuminate hadith processing.

We distinguish several study types rather than treating the literature as one flat pool. Review papers are analyzed as prior syntheses, not as evidence of technical progress. Primary technical and resource papers provide the basis for paper-level appraisal. Conceptual ethics and scholar-perspective papers are used to interpret governance, authority, and acceptable-use questions, but not to support technical performance claims. For close appraisal of primary studies, we selected papers from the core inventory that met four additional criteria: task centrality, methodological distinctiveness, influence on later work or benchmark practice, and sufficient technical detail to support critical evaluation.

\subsection{Appraisal framework}

Rather than rank the literature with a single score, we coded primary studies against explicit dimensions summarized in Table~\ref{tab:appraisal_framework}. This gives the review a structured basis for judgment while respecting the heterogeneity of tasks, datasets, and evaluation settings across the field.

\begin{table}[!htbp]
\caption{Appraisal dimensions used to evaluate primary studies in this review.}
\label{tab:appraisal_framework}
\small
\begin{tabularx}{\textwidth}{p{2.4cm}Y Y}
\toprule
\textbf{Dimension} & \textbf{Question asked} & \textbf{Why it matters} \\
\midrule
Corpus realism & Does the study rely on canonical-only, synthetic, tightly curated, or structurally diverse material? & Separates progress on tidy benchmarks from progress likely to survive real textual variation. \\
\addlinespace
Setting & Is the study evaluated on token labeling, end-to-end workflows, retrieval, expert judgment, or mixed criteria? & Prevents false equivalence between heterogeneous outcome measures. \\
\addlinespace
Transfer & Does the study test cross-collection robustness, real-vs-synthetic transfer, or out-of-distribution behavior? & Indicates whether reported gains travel beyond one dataset or one editorial tradition. \\
\addlinespace
Reproducibility & Are the dataset, code, prompts, annotation guidelines, or enough implementation details available to audit the claim? & Determines whether the result can be reused, stress-tested, or meaningfully compared. \\
\addlinespace
Expert input & Does the study include domain-expert annotation, validation, or error review? & Matters especially in a religious-text domain where benchmark success alone is insufficient. \\
\addlinespace
Scholarly use & Does the task move closer to authentic scholarly workflows, or does it remain a technical proxy? & Distinguishes benchmark advancement from likely scholarly usefulness. \\
\bottomrule
\end{tabularx}
\end{table}

Table~\ref{tab:appraisal_matrix} then applies these dimensions to the primary studies discussed most closely in the paper. We weight peer-reviewed primary studies most heavily when making technical claims. Preprints are retained only when they introduce substantial infrastructures or datasets already shaping later discussion, and we identify them as such when they are cited. Conceptual or ethics papers are treated as interpretive context rather than technical evidence.

Because task definitions, corpora, and metrics vary sharply across the literature, we do not pool headline scores into a single quantitative ranking. We interpret accuracy, F1, expert ratings, retrieval quality, and end-to-end workflow outcomes only within their task contexts. Throughout the paper, we also distinguish three types of claims: empirical observations directly supported by the coded literature, interpretive syntheses across heterogeneous studies, and normative recommendations for future research.

\subsection{Limitations and interpretive stance}

Like any narrative review, ours remains vulnerable to search and selection bias, privileges studies that are indexed or digitally accessible through mainstream scholarly databases, and necessarily underrepresents gray literature, Arabic-only venues with weak indexing, and unpublished tooling used in practice. The long tail of obscure hadith corpora is itself part of the problem: in many cases, discoverable datasets or reproducible computational studies do not yet exist. We therefore present claims about field maturity as a critical synthesis of representative evidence rather than as exhaustive measurement of every relevant project.

Our interpretation is guided by explicit priorities. We give particular weight to provenance, grounding, benchmark realism, and scholar-facing usefulness when reading the literature. That preference reflects the domain under review and the downstream uses we consider especially important, but it also shapes which studies we find especially persuasive. We make that preference explicit because it influences both appraisal and the research agenda proposed later in the paper.

For interpretive clarity, we use a three-level view of the hadith computation pipeline inherited from earlier literature. Level 1 concerns isnad-matn separation and related segmentation tasks. Level 2 concerns narrator analysis, source verification, question answering, and authentication-related modeling. Level 3 concerns knowledge graphs, corpus-wide enrichment, and larger research infrastructures. We use this framework pragmatically rather than doctrinally. It helps explain why progress has been uneven: Level 1 benefited most quickly from clean task definitions and reusable annotations, whereas Levels 2 and 3 depend much more heavily on structured knowledge, cross-document linkage, and scholar-facing validation.
\FloatBarrier

\section{The field before the transformer turn}\label{sec:baseline}

Before the current AI turn, hadith computation already had several recognizable strands: segmentation of isnad and matn \citep{boella2011salah, maraoui2018segmentation}, narrator extraction and graph construction \citep{azmi2010itree, azmi2010enarrator, siddiqui2014extraction, zaraket2012arabic}, classification and authentication support \citep{saloot2016hadith, luthfi2018digital}, and corpus-building efforts for classical Arabic and hadith materials \citep{mahmood2018multilingual, alosaimy2017sunnah}. The field had shown that hadith texts contain exploitable linguistic and structural regularities, especially in canonical collections and well-edited corpora.

The earlier landscape also had clear constraints. Most systems were collection-specific, frequently optimized for highly regular corpora such as Sahih al-Bukhari or Sahih Muslim. Tasks were usually treated in isolation rather than as components of a reusable workflow. Handcrafted rules, trigger words, and local feature engineering remained central even when machine learning was used. Evaluation was fragmented across accuracy, F1, success rate, or ad hoc heuristics, often on incomparable datasets \citep{azmi2019computational}. We emphasize these constraints because many contemporary papers look more transformative than they are once the baseline is forgotten.

We therefore judge the current phase by a harder standard than mere model novelty. The key question is whether later papers reduce the old dependence on narrow corpora, brittle transfer, weak benchmarking, shallow knowledge integration, and limited scholar-facing validation. That is far more demanding than asking whether reported scores went up.

\section{How the field changed in the transformer and LLM era}\label{sec:evolution}

\subsection{The method shift was not a clean handover from old AI to LLMs}\label{sec:method_shift}

\begin{table}[!htbp]
\caption{Main AI paradigms in contemporary hadith computation.}
\label{tab:method_shift}
\small
\begin{tabularx}{\textwidth}{p{2.5cm}Y Y}
\toprule
\textbf{Paradigm} & \textbf{What it demonstrably improved} & \textbf{Why the evidence remains limited} \\
\midrule
Rule-based and hybrid pipelines & Remain competitive on stable segmentation and extraction problems, especially where domain cues are explicit and annotation is limited. & Performance is often brittle outside regular corpora, and many rules encode assumptions that fail on irregular or noisy texts. \\
\addlinespace
Classical supervised models & Provide strong baselines for curated tasks and help show that some gains come from cleaner data rather than architecture novelty. & Results are often collection-specific and depend on handcrafted features or narrow train-test conditions. \\
\addlinespace
Transformer encoders and neural sequence models & Improve contextual modeling for segmentation, retrieval, and labeling tasks and raise the ceiling on multilingual or large-scale processing. & Cross-collection evidence is still thin, and upstream tokenization or segmentation errors can still degrade downstream gains. \\
\addlinespace
Knowledge graphs, ontologies, and retrieval-grounded systems & Move the field toward semantic access, provenance-aware retrieval, and inspectable evidence structures. & Coverage bottlenecks, manual curation burdens, and unresolved entity-linking problems limit claims of broad scholarly reasoning. \\
\addlinespace
LLM-assisted pipelines & Expand corpus-scale enrichment, simplification, multilingual interfaces, and grounded evaluation workflows. & Hallucination risk, cost, prompt sensitivity, incomplete reproducibility, and weak comparability with earlier benchmarks make many claims provisional. \\
\bottomrule
\end{tabularx}
\end{table}

Table~\ref{tab:method_shift} summarizes a distinction that descriptive reviews can understate. Contemporary hadith computation has not progressed through a clean paradigm replacement in which newer models simply superseded older ones. Older symbolic or hybrid methods still provide some of the clearest evidence on tightly defined low-level tasks, whereas transformer, retrieval-grounded, and LLM-based systems have mainly expanded the field's scope, workflow integration, and ambition. This distinction matters because newer architectures should be judged not only by output quality, but also by whether they improve transfer, grounding, auditability, and scholar-facing usefulness under harder conditions \citep{altammami2019hadith, muther2020automatic, kamran2023semantichadith, mubarak2025islamiceval, asgari2025rezwan}.

\begin{table}[!htbp]
\caption{Structured appraisal of representative primary studies. `Y` = yes, `P` = partial, `N` = no.}
\label{tab:appraisal_matrix}
\scriptsize
\begin{tabularx}{\textwidth}{p{2.15cm}p{1.35cm}p{1.7cm}p{0.8cm}p{0.8cm}p{0.85cm}Y}
\toprule
\textbf{Study} & \textbf{Task} & \textbf{Corpus / setting} & \textbf{Xfer} & \textbf{Open} & \textbf{Expert} & \textbf{Main appraisal} \\
\midrule
Altammami et al. \citeyearpar{altammami2019hadith} & Segmentation & Six canonical books; end-to-end benchmark & Y & P & N & Cross-book baseline, but still canonical-bound. \\
\addlinespace
Muther and Smith \citeyearpar{muther2020automatic, muther2020tracing} & Isnad extraction & Long classical texts; ambiguity-aware annotation & P & P & N & More realistic extraction setup; ambiguity remains central. \\
\addlinespace
Abdi et al. \citeyearpar{abdi2020qa} & Hadith QA & Controlled retrieval and QA pipeline & N & P & N & Downstream move toward QA, but in a controlled setting. \\
\addlinespace
Mahmoud et al. \citeyearpar{mahmoud2022arsanad} & Narrator disambig. & Artificial and real sanads; label-heavy classification & P & P & N & Valuable resource, but synthetic-to-real transfer remains weak. \\
\addlinespace
Mghari et al. \citeyearpar{mghari2022sanadset} & Corpus infrastructure & 650K narrations from 926 books & N & P & N & Major diversity gain, but not itself a transfer solution. \\
\addlinespace
Wiharja et al. \citeyearpar{wiharja2022kgqa} and Kamran et al. \citeyearpar{kamran2023semantichadith} & KG QA / retrieval & Bounded KGs and ontology-driven retrieval & N & P & N & Scholar-facing retrieval is promising, but still coverage-bound. \\
\addlinespace
Mubarak et al. \citeyearpar{mubarak2025islamiceval} & Grounded evaluation & Qur'an/Hadith QA and hallucination tasks & N & Y & P & Makes faithfulness testable, but remains a proxy task. \\
\addlinespace
Asgari-Bidhendi et al. \citeyearpar{asgari2025rezwan} & LLM pipeline & 1.2M narrations; multi-task expert scoring & N & N & Y & Infrastructural leap, but reproducibility remains limited. \\
\bottomrule
\end{tabularx}
\end{table}

\subsection{Data and infrastructure became first-class research objects}\label{sec:data}

One of the clearest recent changes is that data work moved from a supporting role to a central research contribution. Earlier papers often treated datasets as local inputs to a model. Contemporary work increasingly treats corpora, annotations, and benchmark assets as contributions in their own right. \citet{altammami2019hadith} and \citet{altammami2020constructing} helped normalize reusable annotated resources for segmentation and bilingual hadith processing. \citet{mghari2022sanadset} pushed scale much further with Sanadset 650K, spanning 926 books and exposing structural variation that canonical-only benchmarking had long obscured. Open infrastructures such as OpenITI \citep{miller2022openiti} also widened the broader text-processing environment in which hadith-specific systems can now operate.

This infrastructural turn changes what we can infer about the field itself. Large and diverse resources make it harder to treat canonical collections as representative of all hadith material. Sanadset shows that substantial portions of large-scale narration data do not fit the structural assumptions built into many earlier experiments \citep{mghari2022sanadset}. This suggests that part of the field's earlier progress was benchmark-relative progress on unusually tidy material.

At the same time, more data did not automatically produce cleaner evidence. \citet{mahmoud2022arsanad} created a valuable narrator-disambiguation resource, but the contrast between validation performance on artificial sanads and weaker results on real test data is instructive. We read that contrast as evidence that large synthetic resources can advance task formalization while still leaving a serious synthetic-to-real transfer gap. Contemporary hadith computation is therefore stronger on data infrastructure than the earlier field, but the evidence for benchmark realism remains limited.

\subsection{Level 1 matured fastest, but the benchmark context still matters}\label{sec:level1}

Within the reviewed literature, the clearest consolidation appears in Level 1 tasks, especially isnad-matn separation and related chain extraction. \citet{altammami2019hadith}, \citet{altammami2020isnad}, \citet{tarmom2020automatic}, and \citet{tarmom2020comparative} show that hybrid methods, compression-based methods, and newer classifiers can all perform strongly when the task environment is stable. We do not read this literature as evidence that one architecture decisively prevailed. Rather, it suggests that segmentation became a more reproducible benchmark area than it had been before.

A more important conceptual advance lies in the move toward more realistic evaluation. \citet{muther2020automatic} and \citet{muther2020tracing} showed that exact boundaries and transmission-chain extraction are not always objectively clean, especially in longer classical Arabic texts. This matters because it changes how performance numbers should be interpreted. A system may identify the correct region while missing a boundary choice on which humans also disagree. In that sense, recent work strengthened the literature not only by raising scores, but also by making the evaluation problem itself less naive.

Even here, however, the field is not finished. Strong segmentation results still depend heavily on corpus structure, and preprocessing quality remains a bottleneck. \citet{alshuhayeb2025noor} show that hadith-domain Arabic word segmentation remains substantially harder than many general NLP pipelines assume. This is a useful reminder that LLM-era optimism can obscure low-level linguistic fragility. If tokenization and morphological segmentation are weak, downstream gains at Level 2 will remain unstable no matter how impressive the larger model appears.

\subsection{Level 2 broadened beyond classification, but progress remains uneven}\label{sec:level2}

One clear Level 2 change is breadth. Hadith computation is no longer framed only around segmentation or binary authenticity classification. It now includes narrator disambiguation \citep{mahmoud2022arsanad}, source-verification tasks \citep{syed2019verifying}, question answering \citep{abdi2020qa, wiharja2022kgqa}, and knowledge representation \citep{kamran2023semantichadith}. We regard that expansion as meaningful progress.

That breadth should not be conflated with maturity. Several representative studies solve a well-defined subproblem while leaving the hardest hadith-science bottlenecks in place. \citet{syed2019verifying} is important because it moves the field closer to philological error detection and citation verification rather than generic text classification. It remains, however, focused on a specific corrective problem rather than a full authenticity framework. \citet{abdi2020qa} shows that hadith question answering can benefit from linguistic knowledge, but the system operates in a more controlled environment than live scholarly retrieval. \citet{wiharja2022kgqa} and \citet{kamran2023semantichadith} push the field toward graph-based semantic access, yet they remain bounded by the coverage and quality of the knowledge bases they build.

We would therefore not describe Level 2 as ``solved'' or even clearly convergent. A more accurate description is that it is diversifying under pressure. The field has more tasks, more modeling strategies, and better formalization than before, but it remains weak on cross-collection robustness, entity linking across biographical traditions, and benchmark standards that would support confident claims of general progress.

\subsection{Level 3 shifted from isolated structures to pipeline-scale enrichment}\label{sec:level3}

A major Level 3 change is the field's move toward larger infrastructures rather than isolated tools. Earlier graph-building and cross-document work \citep{azmi2010enarrator, zaraket2012arabic} showed what was possible, but those efforts were still local in scale. By contrast, contemporary work increasingly frames hadith computation as an end-to-end environment involving OCR repair, segmentation, validation, semantic tagging, knowledge graph construction, multilingual access, and expert review.

\citet{asgari2025rezwan} is among the clearest expressions of that change. Its importance is not only technical. The paper broadens the field's ambition by treating large-scale hadith enrichment as a research infrastructure problem rather than a single-task benchmark. At the same time, the emergence of grounded evaluation frameworks such as \citet{mubarak2025islamiceval} indicates that the LLM era is also pushing the field to examine hallucination, source faithfulness, and reliable retrieval more explicitly.

Recent work points in the same direction. \citet{abbas2026fanarsadiq} move toward grounded multi-agent Islamic question answering rather than unconstrained generation. The literature therefore suggests a shift not toward purely generative systems, but toward more pipeline-oriented and grounded ones. That distinction matters because hadith computation gains little from fluent outputs that cannot be traced back to authoritative evidence.
\FloatBarrier

\section{Critical appraisal of representative contemporary studies}\label{sec:appraisal}

We still need a review of this kind because recent original studies do not all contribute in the same way. Some expose structural realities the field needed to confront. Others produce useful resources. Others solve only proxies of the problems they are taken to represent. Four patterns stand out in the reviewed sample.

First, several of the strongest papers improved the field by making benchmarks more realistic rather than merely by raising scores. \citet{muther2020automatic} and \citet{muther2020tracing} matter because they confront ambiguity in extraction. \citet{mghari2022sanadset} matters because it exposes the structural diversity of real-world narration data. These papers are methodologically valuable because they make optimistic simplifications harder to maintain.

Second, some high-value papers reveal how far the field still is from real scholarly deployment. \citet{mahmoud2022arsanad} is an important example. It formalizes narrator disambiguation at scale and provides a usable benchmark resource, but the gap between artificial-sanad validation and real-test performance is exactly the kind of evidence a descriptive review can miss. The task should not be treated as solved simply because it is now trainable.

Third, contemporary hadith QA and knowledge-graph papers are promising but should be read carefully. \citet{abdi2020qa}, \citet{wiharja2022kgqa}, and \citet{kamran2023semantichadith} all push the field toward scholar-facing interfaces and semantic organization. That is progress. But these systems still depend heavily on controlled corpora, bounded ontologies, or limited retrieval setups. They are better read as enabling infrastructures than as substitutes for hadith-critical reasoning.

Fourth, LLM-era papers widen the field's horizon while introducing new evidentiary problems. \citet{altammami2025leveraging} is notable because it uses an Islamic Studies expert to verify simplification outputs, making accessibility work methodologically serious rather than merely generative. \citet{mubarak2025islamiceval} is valuable because it turns faithfulness and correction into explicit evaluation targets. \citet{asgari2025rezwan} shows what large-scale enrichment can look like with expert scoring and multilingual layers. Yet these LLM-era contributions are also harder to compare with earlier work. Expert ratings, task bundles, cost analyses, and corpus-scale pipelines are useful, but they are not directly commensurable with the classic segmentation benchmarks that dominated the pre-LLM field. A review that does not confront this comparability problem is likely to overstate coherence.

Across these categories, the same reviewer-level concerns recur. Too many papers rely on narrow corpora, compare against weak or heterogeneous baselines, report limited error analysis, or infer broad scholarly usefulness from what are still bounded technical proxies. That does not nullify the contributions. It does mean the field needs tougher standards of evidence than it often imposes on itself.

Taken together, the recent original literature points to a field that is improving in three meaningful ways: it is using larger resources, asking more realistic questions, and taking grounding more seriously. It also points to persistent weaknesses: benchmark insularity, proxy-task drift, synthetic-to-real transfer gaps, and limited reproducibility for the most ambitious LLM systems. This is why we regard paper-level critique as necessary rather than optional.

\section{Major gaps still defining hadith computational science}\label{sec:gaps}

\subsection{The field remains overly concentrated on the six canonical books}\label{sec:gap_corpora}

One of the most important structural gaps is corpus concentration. Much of the field's strongest benchmark culture still grows out of the six canonical Sunni collections, or even narrower subsets of them \citep{altammami2019hadith, altammami2020constructing, abdi2020qa}. This focus is understandable. These collections are well edited, widely digitized, and structurally regular enough to make segmentation and retrieval experiments tractable. But the same convenience has shaped the field's blind spots.

Hadith scholarship is much larger than the six canonical books. It includes musnads (collections organized primarily by narrator), musannafs (topic-organized compilations), mu'jams (collections arranged by names or teachers), ajza' (small booklet-style collections), later compilations, regional collections, rijal works (narrator-biographical literature), takhrij literature, sectarian corpora, and many partially digitized or poorly OCR'd texts that remain outside mainstream benchmarks. \citet{mghari2022sanadset} help expose this problem by scaling to 926 books, and \citet{asgari2025rezwan} show that larger and more diverse repositories are now technically processable. Even so, the benchmark culture of the field still lags behind the textual reality of the tradition. From the literature we reviewed, we infer that there is still no comparable shared-benchmark ecosystem for the long tail of hadith literature, especially obscure, regionally transmitted, or manuscript-adjacent works.

This gap is not only about fairness to neglected texts. It is also about scientific validity. A field that learns mainly from structurally regular canonical collections risks overestimating transferability, underestimating OCR and metadata problems, and confusing editorial cleanliness with true task maturity. Future research should therefore treat long-tail corpus development as a first-order scientific objective. That means collection-aware metadata, edition tracking, OCR benchmarking for Arabic religious texts \citep{heakl2025kitabbench}, and shared tasks that explicitly include irregular, incomplete, and obscure material rather than treating it as noise.

\subsection{The explanatory layer of hadith scholarship is still largely missing}\label{sec:gap_commentary}

Another major gap is the relative absence of computational work on hadith explanation rather than hadith text alone. Most hadith NLP papers stop at segmentation, narrator processing, classification, retrieval, or source verification. They rarely move into the commentarial layer in which hadith meaning is clarified, variant reports are reconciled, legal implications are debated, and lexical or contextual difficulties are resolved. From the literature assembled for this review, we infer that direct computational treatment of sharh al-hadith (hadith commentary), takhrij reasoning, and fiqh al-hadith (legal and interpretive analysis of hadith) remains sparse relative to work on base hadith text.

This gap matters because hadith scholarship is not reducible to the bare matn plus isnad. Scholars often rely on commentaries, cross-references, gradings, sabab al-wurud (occasion or circumstance of narration) discussions, and juristic interpretation to decide what a narration means and how it should be used. The nearest adjacent progress has happened on the Qur'anic side, where question answering and retrieval resources are beginning to connect text with tafsir (Qur'anic exegesis) \citep{alnefaie2023haqa, alazani2025ontologyragq}. Hadith computation has not yet built comparable resources for major commentaries or explanatory traditions.

We regard this as an important future direction. The field needs aligned datasets that connect base narrations to commentary spans, explanatory glosses, grading arguments, and juristic inferences. It also needs models that can distinguish between the base hadith, the commentator's paraphrase, the legal inference drawn from it, and the school-specific limitations placed on that inference. Without that layer, hadith computation will remain text-processing-heavy but scholarship-light.

\subsection{Hadith is still weakly connected to the wider Islamic knowledge network}\label{sec:gap_integration}

Hadith belongs to a larger Islamic scholarly system. Its interpretation often depends on Qur'anic context, prophetic biography (seerah), narrator biography, tafsir, legal chapters, and later juristic synthesis. Yet most computational work still treats hadith as an isolated text collection. In our view, that isolation is increasingly out of step with how Islamic scholarship actually works.

There are promising adjacent efforts. \citet{altammami2021jointontology}, \citet{altammami2022quranhadith}, and \citet{alshammari2024linking} show that linking Qur'an and Hadith is computationally feasible. \citet{alnefaie2023haqa} and \citet{alazani2025ontologyragq} illustrate how grounded question answering and retrieval can be built around Qur'anic materials. \citet{nakhlah2023qasina} shows that prophetic biography can also enter the computational question-answering space. But these efforts remain mostly pairwise, task-specific, or adjacent to hadith computation rather than integrated into its center.

We therefore locate the deeper gap not in the total absence of cross-source work, but in the absence of a mature multi-hop research framework that links verse, hadith, seerah event, commentary, fiqh chapter, narrator biography, and later scholarly usage in one inspectable evidence graph. Such a framework would better reflect how Islamic scholarship derives meaning and rulings. It would also change what hadith computation systems are optimized for: not isolated classification, but contextualized evidence navigation.

\subsection{The bridge from hadith computation to contemporary fiqh remains underbuilt}\label{sec:gap_fiqh}

The final gap is especially consequential for present-day use. Hadith computation has obvious relevance for contemporary fiqh, but the connection is still weakly developed in the literature. Emerging systems in Islamic QA, inheritance reasoning, and fatwa generation show that the applied jurisprudential layer is already moving computationally \citep{alyemny2023fatwaset, bouchekif2025qias, mohammed2025aftina, abbas2026fanarsadiq}. Yet hadith computation research is only partially connected to that movement.

This disconnect matters because fiqh rarely depends on hadith in isolation. It depends on evidence selection, authenticity assessment, reconciliation of apparently conflicting narrations, juristic interpretation, school-specific methodological filters, and linkage to Qur'anic and contextual evidence. A hadith-processing system that returns a single narration without provenance, commentary, variants, or juristic framing is not yet a serious fiqh support tool.

We therefore argue that the proper role of hadith computation in contemporary fiqh is not autonomous fatwa issuance, but evidence support. Future systems should help scholars, researchers, and advanced students retrieve relevant narrations, surface parallel or variant reports, connect them to Qur'anic and seerah context, expose relevant commentary and takhrij, show where madhhab-specific usage diverges, and present uncertainty rather than suppress it. That would make hadith computation genuinely useful to present-day jurisprudential reasoning while respecting the limits of automation in a normatively sensitive field. Here, madhhab refers to a legal school with its own methodological filters for weighing evidence.

\begin{figure}[!htbp]
\centering
\resizebox{\textwidth}{!}{%
\begin{tikzpicture}[
  font=\small,
  node distance=0.8cm and 0.8cm,
  source/.style={draw, rounded corners, align=center, minimum width=2.4cm, minimum height=0.95cm, fill=gray!10},
  core/.style={draw, rounded corners, align=center, minimum width=10.4cm, minimum height=1.35cm, fill=blue!8},
  outcomebox/.style={draw, rounded corners, align=center, minimum width=10.4cm, minimum height=1.15cm, fill=green!8},
  edge/.style={-{Latex[length=2.2mm]}, thick}
]
\node[source] (quran) {Qur'an\\and tafsir};
\node[source, right=of quran] (hadith) {Hadith corpora\\canonical + long-tail};
\node[source, right=of hadith] (sharh) {Sharh, takhrij,\\fiqh al-hadith};
\node[source, right=of sharh] (seerah) {Seerah and\\historical context};
\node[source, right=of seerah] (rijal) {Rijal and\\narrator biography};

\node[core, below=1.15cm of hadith, xshift=2.0cm] (core) {Integrated hadith computational science\\
\footnotesize OCR and digitization, segmentation, entity linking, retrieval, knowledge-graph alignment, grounding, expert validation};

\node[outcomebox, below=1.05cm of core] (outcome) {Scholar-facing outcomes\\
\footnotesize contextual understanding, provenance-aware QA, disagreement mapping, evidence support for contemporary fiqh};

\draw[edge] (quran.south) -- (core.north west);
\draw[edge] (hadith.south) -- (core.north);
\draw[edge] (sharh.south) -- (core.north);
\draw[edge] (seerah.south) -- (core.north east);
\draw[edge] (rijal.south) -- (core.north east);
\draw[edge] (core.south) -- (outcome.north);
\end{tikzpicture}
}
\caption{Conceptual foundation for the next phase of hadith computational science. The field should move from isolated hadith-text tasks toward an integrated evidence layer that connects hadith with Qur'an, commentary, seerah, and biographical scholarship in support of inspectable scholarly and fiqh-facing workflows.}
\label{fig:integrated_pipeline}
\end{figure}
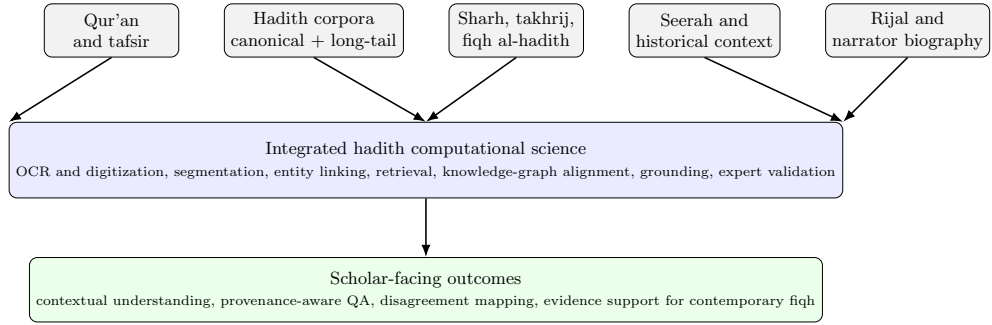
\FloatBarrier

\section{Islamic scholar and domain-expert perspectives}\label{sec:scholar_perspective}

Recent review papers still leave an important gap because they rarely integrate the perspective of Islamic scholars and domain experts in a substantive way. That omission matters because the key questions in hadith computation are not purely technical. They involve authenticity, interpretive authority, and the acceptable role of automation in a tradition where evidentiary discipline is central.

\citet{akbar2024digitalturn} describe the contemporary digital turn in hadith studies as a mixed development: digital platforms expand access, education, and discoverability, but they also accelerate the circulation of unverified narrations, intensify algorithm-driven visibility, and weaken the gatekeeping role of trained scholars. This argument matters for hadith computation because it changes the standard for useful AI. In this domain, a system is not valuable simply because it retrieves or generates more text. Its value also depends on whether it preserves chains of authority, provenance, and critical scrutiny.

\citet{abdulrahman2024future} make a related point from a more programmatic angle. Their discussion of opportunities and challenges emphasizes that digital hadith work should be judged by how well it integrates technological efficiency with standards of trust, authenticity, and honesty. The paper is less methodologically sharp than core computational studies, but it captures an important scholarly concern: the relevant question is not only whether AI can process hadith material, but also under what constraints it should do so.

The broader Islamic AI ethics literature strengthens this concern. \citet{elmahjub2023islamicethics} argues for pluralist ethical benchmarking rather than purely technical optimization, while \citet{nawi2021muslimexperts} report that Muslim experts see a strong need for regulation and frameworks grounded in maqasid al-shari'ah, that is, the higher objectives of Islamic law. Although these are not hadith-specific papers, they help explain why expert-aligned evaluation is not an optional extra in a religious text domain. They also show why performance-only review papers can feel incomplete: they leave aside the governance question that increasingly shapes whether an AI system is acceptable in practice.

At the same time, the available scholar-perspective literature is not as broad as the field sometimes implies. Much of it is conceptual, normative, or based on small expert pools rather than broad comparative evidence across madhhabs, that is, legal schools, institutions, and levels of technical literacy. The current evidence therefore does not justify treating any single paper as a proxy for ``the'' Islamic scholar view. What it does support is narrower but still important: qualified domain experts repeatedly demand provenance, transparency, uncertainty awareness, and human oversight, while the field still lacks standardized protocols for eliciting and evaluating scholar judgment inside AI studies \citep{nawi2021muslimexperts, akbar2024digitalturn, abdulrahman2024future}.

Recent technical work is beginning to absorb this lesson. \citet{altammami2025leveraging} explicitly verifies simplification outputs with an Islamic Studies expert. \citet{asgari2025rezwan} evaluate a large LLM-assisted pipeline using six domain experts rather than relying only on automatic metrics. \citet{mubarak2025islamiceval} define grounded tasks against authoritative Qur'an and Hadith sources, making hallucination and correction central evaluation concerns. Taken together, these studies suggest a gradual shift from model-centric claims toward workflows in which experts, sources, and grounding matter.

This scholar perspective also helps clarify current trends. A notable direction in the recent literature is the turn away from generic ``Islamic chatbot'' development and toward grounded systems that retrieve, verify, or correct religious content rather than improvise it. In that sense, newer systems such as \citet{abbas2026fanarsadiq} are significant less because they are agentic, and more because they treat groundedness as a design principle. That is closer to what hadith scholarship demands.

\section{What the LLM era actually changed}\label{sec:llm_turn}

Current AI discourse often implies a simple narrative: old NLP pipelines gave way to LLMs, and serious language tasks should now be understood through that lens alone. We do not think hadith computation fully fits that story. LLMs did change the field, but mainly by changing ambition, workflow design, and the burden of proof.

First, LLMs changed scale. Systems such as Rezwan make it plausible to process hundreds of thousands or millions of narrations with layered enrichment, multilingual access, and expert review \citep{asgari2025rezwan}. That would have been impractical under the old craft model of local pipelines built around small corpora.

Second, LLMs changed workflow design. The unit of innovation is no longer always the standalone classifier. Increasingly, the field works with orchestrated pipelines that include OCR repair, segmentation, retrieval, source checking, simplification, semantic tagging, and human validation. This is visible both in large enrichment pipelines and in grounded evaluation tasks \citep{mubarak2025islamiceval, abbas2026fanarsadiq}.

Third, LLMs changed what counts as convincing evidence. In the pre-LLM era, a strong number on a curated corpus could often stand as the paper's main claim. In the current phase, that is less persuasive by itself. Readers want to know how a system behaves on structurally diverse data, whether outputs can be grounded in authoritative sources, whether experts judge the errors tolerable, and whether the workflow is reproducible enough to be trusted.

We should be equally clear about what LLMs did not change. They did not solve narrator identity resolution. They did not make preprocessing irrelevant. They did not eliminate the need for explicit biographical and bibliographic grounding. And they did not erase the difference between fluent text generation and reliable hadith scholarship. For that reason, we describe the current landscape as hybrid rather than purely LLM-centered. The most credible systems in this field are likely to combine structured resources, retrieval, task-specific models, and expert supervision rather than rely on unconstrained generation alone.

\section{A research agenda for the next phase}\label{sec:agenda}

Building on earlier surveys, we argue that the next phase should be organized around a smaller number of hard problems and clearer scholarly standards. In our view, the agenda is best understood as a set of linked research programs rather than as a list of isolated recommendations.

\subsection{Benchmark reform and evaluation realism}

The first program concerns evaluation itself. Shared benchmarks need to move beyond canonical within-collection testing toward cross-collection transfer, structurally irregular narrations, obscure collections, and manuscript- or OCR-contaminated material. Without that shift, the field will continue to confuse performance on tidy editorial settings with performance on the wider hadith tradition. This program also requires clearer reporting standards: future studies should state corpus boundaries, preprocessing assumptions, openness of data and code, cross-domain testing strategy, and failure modes in enough detail for independent stress-testing.

\subsection{Knowledge infrastructure and long-tail corpus building}

The second program concerns resource creation. The field needs a long-tail corpus effort that extends beyond the six canonical books to musnads, musannafs, mu'jams, rijal works, commentaries, and other under-studied corpora. Scientifically, this matters because transfer claims remain weak without broader textual coverage. Practically, it matters because many of the texts that shape scholarship are still poorly digitized, inconsistently edited, or difficult to align across editions. A credible infrastructure program should therefore include collection-aware metadata, edition tracking, OCR benchmarking, reusable annotation guidelines, and explicit citation practices that treat datasets and corpora as research assets rather than incidental inputs.

\subsection{Scholarship-aware modeling}

The third program concerns task design. Narrator identity resolution still needs explicit biographical infrastructure, including well-linked rijal resources, entity-linking datasets, and temporal or geographic constraints. At the same time, commentary-aware hadith computation should become a substantive research front rather than a peripheral aspiration. The field needs datasets and models that align narrations with sharh, takhrij, lexical explanation, and juristic inference, and that distinguish between the base report, later commentary, and school-specific interpretation. Progress here would move the literature away from proxy-task accumulation and toward workflows closer to actual scholarly practice.

\subsection{Integrated Islamic knowledge systems and fiqh-facing support}

The fourth program concerns integration across sources. Future systems should connect hadith with Qur'an, seerah, tafsir, fiqh chapters, and narrator biographies in inspectable evidence networks rather than operate as isolated text silos. This is important not only for retrieval quality, but also for contextual fidelity. In contemporary fiqh-facing settings, the relevant need is seldom decontextualized retrieval; it is provenance-rich evidence support that can surface parallel reports, commentary, disagreement, and juristic framing. That goal does not imply automated fatwa issuance. It implies better computational assistance for scholars, researchers, and advanced students working within an existing evidentiary tradition.

\subsection{Expert-centered governance and evaluation}

The fifth program concerns governance and acceptable use. LLM systems in this area should be evaluated for grounding, cost, reproducibility, uncertainty, and expert inspection, not only for fluency. In practice, that means scholar-in-the-loop evaluation protocols, auditable prompts or workflow descriptions, transparent handling of authoritative sources, and clearer boundaries around what a system is and is not designed to do. In a religious-text domain, provenance, faithful citation, uncertainty reporting, and the role of qualified scholarship are part of system quality itself, not peripheral reflection.

Taken together, these programs imply a broader principle. We argue that hadith computation should define success in terms appropriate to its own domain rather than borrow legitimacy from general AI enthusiasm. The field is likely to progress most when it treats benchmark realism, long-tail corpus coverage, commentary awareness, knowledge integration, and collaboration with hadith and fiqh scholars as integral to technical design rather than as afterthoughts.

\section{Conclusion}\label{sec:conclusion}

Recent review papers have mapped publication growth and topical trends, but they do not fully explain the field's present trajectory. Our reading of the reviewed literature suggests that the clearest recent advances lie in data infrastructure, more realistic segmentation and extraction work, better-formalized narrator and source-verification tasks, and the emergence of grounded LLM-era pipelines. The most persistent weaknesses remain cross-collection robustness, narrator identity resolution, preprocessing, benchmark comparability, reproducibility, and epistemic grounding.

Our contribution is threefold. We critically examine the current review literature rather than merely citing it. We evaluate representative original papers at the level of methodological strength and limitation. And we show that Islamic scholar and domain-expert perspectives are now central to understanding the field's trajectory, not external commentary that can be appended later.

Our main conclusion is strategic but provisional. We do not see hadith computation as transformed by a simple replacement of older methods with LLMs. We see it as having entered a hybrid phase in which structured resources, neural models, retrieval, knowledge graphs, and expert supervision all matter. We therefore treat many current headline gains as promising but still provisional, especially where they depend on narrow corpora, synthetic data, weak baselines, or incomparable evaluation setups. In our reading, the next advances are most likely to come from better grounding, better benchmarks, broader corpus coverage beyond the canonical six, computational engagement with commentaries and explanatory traditions, and tighter integration between AI practice and the wider ecosystem of hadith, Qur'an, seerah, and fiqh scholarship.

\backmatter

\section*{Glossary of key terms}

\begin{tabularx}{\textwidth}{p{3cm}Y}
\toprule
\textbf{Term} & \textbf{Working meaning in this review} \\
\midrule
Isnad / sanad & Chain of transmission for a hadith report. \\
\addlinespace
Matn & The content or wording of the report itself. \\
\addlinespace
Sharh & Commentary that explains wording, context, or interpretation. \\
\addlinespace
Takhrij & Source tracing and authentication referencing across collections and transmissions. \\
\addlinespace
Rijal & Narrator-biographical literature used to identify and assess transmitters. \\
\addlinespace
Fiqh al-hadith & Legal and interpretive analysis derived from hadith. \\
\addlinespace
Sabab al-wurud & The occasion or circumstance associated with a narration. \\
\addlinespace
Madhhab & A legal school within Islamic jurisprudence. \\
\addlinespace
Maqasid al-shari'ah & The higher objectives of Islamic law. \\
\addlinespace
Musnad & A collection organized primarily by narrator, often by Companion. \\
\addlinespace
Musannaf & A hadith compilation organized by topical or legal chapters. \\
\addlinespace
Mu'jam & A collection arranged by names, teachers, or alphabetical order. \\
\addlinespace
Ajza' & Small booklet-style collections, often focused on one topic or transmitter. \\
\addlinespace
Seerah & Prophetic biography. \\
\addlinespace
Tafsir & Qur'anic exegesis or commentary. \\
\bottomrule
\end{tabularx}

\section*{Declarations}

\textbf{Funding} No external funding was received for this study.

\textbf{Conflict of interest} The authors declare no conflict of interest.

\textbf{Ethics approval and consent to participate} Not applicable.

\textbf{Consent for publication} Not applicable.

\textbf{Data availability} This study is a literature-based narrative review and does not report a new dataset.

\textbf{Materials availability} Not applicable.

\textbf{Code availability} Not applicable.

\textbf{Author contribution} Md. Ashraful Haque led the literature collection, synthesis, and drafting. Riasat Islam contributed to the conceptual framing, critical interpretation, and revision of the manuscript. Both authors approved the final manuscript.

\end{document}